# SPOT-THE-CAMEL: COMPUTER VISION FOR SAFER ROADS

Khalid AlNujaidi, Ghadah AlHabib, and Abdulaziz AlOdhieb

Computer Science Department, Prince Mohammad Bin Fahd University,
Khobar, Saudi Arabia.

## ABSTRACT

*As the population grows and more land is being used for urbanization, ecosystems are disrupted by our roads and cars. This expansion of infrastructure cuts through wildlife territories, leading to many instances of Wildlife-Vehicle Collision (WVC). These instances of WVC are a global issue that is having a global socio-economic impact, resulting in billions of dollars in property damage and, at times, fatalities for vehicle occupants. In Saudi Arabia, this issue is similar, with instances of Camel-Vehicle Collision (CVC) being particularly deadly due to the large size of camels, which results in a 25% fatality rate [1]. The focus of this work is to test different object detection models on the task of detecting camels on the road. The Deep Learning (DL) object detection models used in the experiments are: CenterNet, Efficient Det, Faster R-CNN, SSD, and YOLOv8. Results of the experiments show that YOLOv8 performed the best in terms of accuracy and was the most efficient in training. In the future, the plan is to expand on this work by developing a system to make countryside roads safer.*

## 1. INTRODUCTION

Wildlife-vehicle collisions (WVCs) are a global issue posing significant threats to human safety and wildlife populations. They can cause injuries and fatalities to drivers and passengers and disrupt migration patterns and breeding habits. WVCs occur similarly across continents, involving different species such as deer in North America and Europe, kangaroos in Australia, and camels in the Middle East and North Africa (MENA). The frequency of WVCs has risen over the past century due to human population growth, urbanization, and new road construction and is expected to continue to increase.

WVCs cause various losses such as property damage, ecosystem disturbance, and fatalities. In the US, 247,000 deer-related WVCs occur annually, causing 200 fatalities and $1.1bn in property damage [2]. Larger animals such as moose result in more fatalities for vehicle occupants; in Sweden, 4092 moose-related WVCs occurred in a year with a 5% fatality rate and significant property damage [3]. Camel-Vehicle Collisions (CVCs) are among the most fatal WVCs due to the animal's physical nature, often resulting in the animal falling through the windshield and causing up to 25% fatalities [1]. 22,897 CVCs were recorded from 2015-2018 [4]. Unfortunately, comprehensive data on CVC frequency, location, and property damage is not easily accessible. However, news articles about recent CVCs with graphic details can be found through a simple web search.

There has been, and continues to be, effort put into deploying countermeasures to reduce WVC. The most commonly deployed tactics currently in place are conventional ones such as fencing and reflective warning signs. As effective as they may have been, signs can go unnoticed by drivers, and animals have found ways through placed fences [5]. These methods require significant funds and labour to set up and maintain. Therefore, it is necessary to consider developing smarter, technologically advanced, and autonomous methods to act as countermeasures to WVC. Just as it has become very common around the world to use sensors and computer vision technologies to assist in the enforcement of traffic violations, the same can be done to mitigate WVCs and, as the focus of this work, CVCs.

The focus of this work will be to evaluate different state-of-the-art object detection algorithms to serve as a base for a CVC avoidance system. The vision of the work is to further develop an autonomous mechanism that makes countryside roads safer. As has been covered so far in the introduction, the issue of animals colliding with vehicles is a global issue. There are solutions, though they can be very costly and can be improved upon to utilize newer technologies to achieve better results. In the next section, there will be a review of literature related to solutions for both global WVCs and local CVCs. Furthermore, a section will be dedicated to the discussion of our proposed system and the methodologies used. Lastly, there will be a section to review the results, summarize the work, and express the vision for future development.

## 2. RELATED WORK

The issue of CVC has been present in the MENA region for a long time and researchers have been exploring various solutions. One proposal involves using an omni-directional radar that detects movement and uploads it to the cloud for analysis. If the speed of movement falls within a specific range, alarms will go off [6]. Another proposal involves implanting chips in camels that can be tracked through LoRaWAN and GPS. By monitoring the geographical location of the camels, warnings are sent when they approach roads [7,8]. In [9], a system consisting of night vision cameras and cellular network handsets was proposed to detect and simulate the movement of camels. A review of several animal-vehicle collision methods was conducted in [10], which found that fencing is currently the most effective solution. However, the paper also highlighted the harm and disruption to the ecosystem caused by fencing, and proposed automated gates that open based on proximity of radio signals sent by collars on the camels. In [11], the authors proposed a road-based system consisting of wireless IR sensors connected to a sink node. The sensors are arranged in clusters along the sides of the road and each node is equipped with a thermal camera and ultrasonic sensor. When movement is detected, the sink node takes an image with the thermal camera and analyses the image. If confirmed remotely, alarms will go off.

All of the methods related to the CVC issue rely on human intervention, such as cooperation from the owners of the camels, or require connectivity to the cloud or network, which may not be possible in some remote areas.

Computer vision methods have been proposed to identify wildlife autonomously and mitigate WVC. In Australia, a region-based convolutional neural network solution was proposed in [12] for detecting kangaroos in traffic through a vehicle-based framework that warns drivers of oncoming kangaroos. The system is deployed in vehicles through dashcam-like cameras or 3D LIDAR. The effectiveness of computer vision in automating wildlife detection is demonstrated in [13], where an end-to-end framework was developed for automating image processing and animal detection through an IoT system in a wildlife reserve. If sensors detect movement, an image is captured and sent to a server to classify the animal and add its details to a database. [14] evaluated the use of deep CNNs to track 20 different species, achieving high accuracies of up to 91.4%.

## 3. METHODS

This section will cover the methodology for the experiments. We are taking a computer vision approach to develop a CVC mitigation system. This section will cover computer vision as a technology and the different subtasks used in the experiments. The focus of the experiments is to evaluate different object detection algorithms, which will be explained. Lastly, the details of the dataset will be briefly discussed.

### 3.1. Computer Vision

Computer vision is a field of study aimed at enabling machines to interpret and understand visual information from the world, such as images and videos. It involves developing algorithms and models to analyse and understand digital images and videos and is applied in a wide range of fields, including image classification, object recognition, image retrieval, autonomous vehicles, and medical imaging. It combines computer science, mathematics, and engineering techniques to create systems capable of perceiving, analysing, and understanding visual information [15].

#### 3.1.1. Classifying Images

Image classification is the task of assigning a label or class to an input image based on its visual content. It's a fundamental problem in computer vision and has numerous applications, including object detection.

The image classification process usually involves extracting features from the image that are relevant to the task, and then using these features to classify the image into one of several predefined classes. Convolutional Neural Networks (CNNs), a type of deep neural network, are the most commonly used method for image classification. They have demonstrated high effectiveness, achieving high accuracy in various image classification tasks.

The image classification process begins with a dataset of labelled images, each associated with a class label. The model learns the mapping between image features and class labels by training on this dataset. After training, it can classify new images by passing the input image through the network and determining the class predictions.

CNNs are more suitable for image classification tasks than traditional neural networks for a few reasons [16], primarily due to parameter efficiency. CNNs use fewer parameters to represent the same amount of information, making them more parameter efficient than traditional neural networks.

The main components of CNNs are convolutional and pooling layers. Convolutional layers apply a small number of filters to the input image and move the filters across the image to detect features at different locations, generating feature maps. Pooling layers then follow, reducing the number of parameters in the network, making it more resilient to small translations of the input image, and decreasing the dimensionality of the feature maps. (Figure 1)

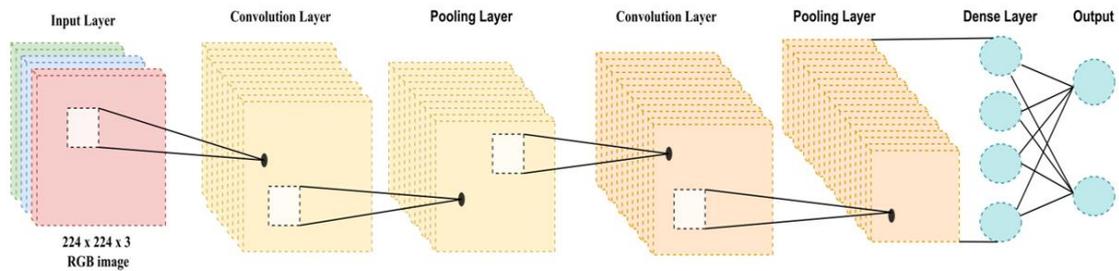

Figure 1. Layers of a basic CNN

Throughout the last decade, researchers have developed different architectures, each of which uses a different number of convolution layers (known as depth), a different number of filters per convolution (known as width), and different mathematical operations within the network (such as the residual blocks introduced in ResNet). These developments have been crucial in combating the vanishing gradient problem that occurs when networks become very deep. Some popular CNN architectures include LeNet, AlexNet, VGGNet, Inception, ResNet, and DenseNet [17, 18, 19, 20, 21, 22]. (Figure 2)

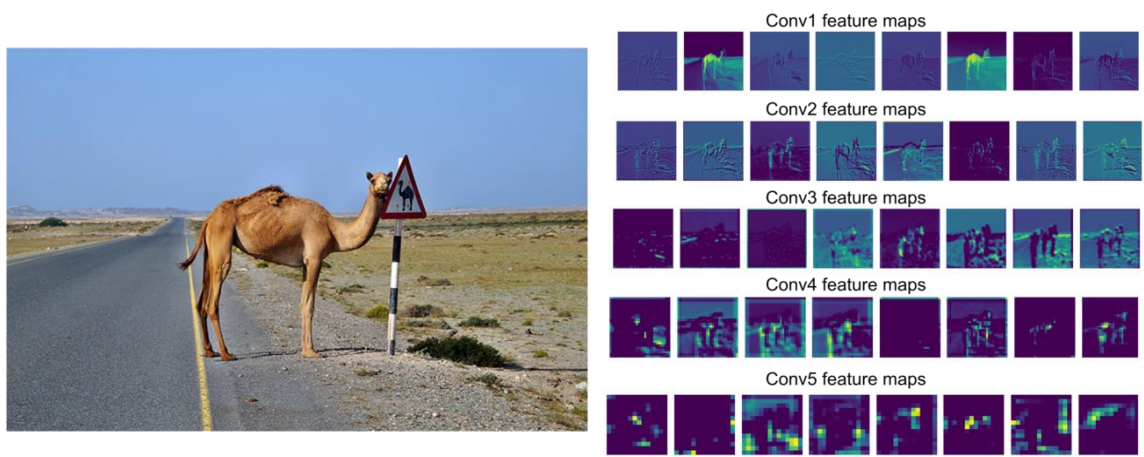

Figure 2. Example outputs of the first 5 convolution layers of VGGNet

### 3.1.2. Detecting Objects

Object detection deals with identifying specific instances of objects of a particular class, such as a camel, in digital images and videos. The development of deep learning has greatly improved object detection, making it capable of detecting objects in images and videos with increased accuracy, efficiency, and real-time capabilities. It's important to note that image classification only recognizes what is present in the image and does not take into consideration the location of the object. Conversely, object detection not only recognizes the class of the object, but also locates it within the image or video. A visualization of the difference can be seen in (Figure 3).

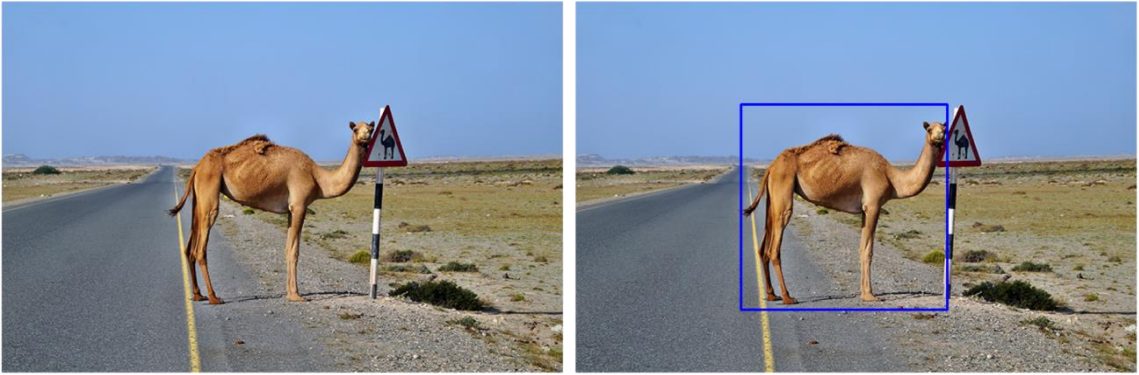

Figure 3. Image on left can be classified having a camel, though the image on right can identify where the camel is exactly.

The process of object detection typically involves two steps. First, localizing the object within the image, typically by identifying its location using bounding boxes (BBs). Secondly, identifying the object's class, such as whether it is a person, car, or camel. CNNs form the backbone of most object detection algorithms due to their remarkable ability to contextualize images.

There are mainly two approaches for object detection algorithms [23]: two-stage methods and one-stage methods. The two-stage method uses a region proposal algorithm to generate "regions of interest" (RoIs) as potential object locations. The classification network then predicts the class and location of each object within these RoIs. However, this approach is computationally expensive as it requires the classification network to be applied to a large number of RoIs. In contrast, the one-stage method divides the image into a fixed grid and applies a classification network to each segment. Figure 3 illustrates the difference between the two approaches, with the number of grids and proposed RoIs greatly reduced for clarity.

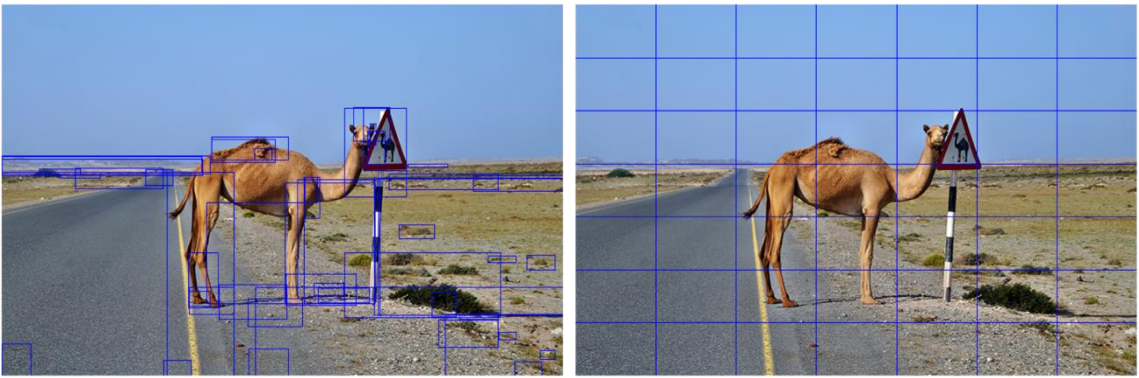

Figure 4. Two-stage approach vs. One-stage approach respectively.

In these experiments, five object detection algorithms are used, with Faster R-CNN being a two-stage detector, and the others (CenterNet, SSD, EfficientDet, and Yolov8) being one-stage detectors.

### 3.2. Dataset Details

The dataset consists of 250 images of camels (Figure 5) in various contexts, such as deserts, captive environments, and highways. These images were gathered from various online sources.

For the use of the object detection algorithms in these experiments the images required annotations. These annotations are BBs of the locations of the object of interest (i.e., camels) within the image to assist in training a model based on the used algorithms. A handy tool named 'Open Labelling' [24] annotating the images.

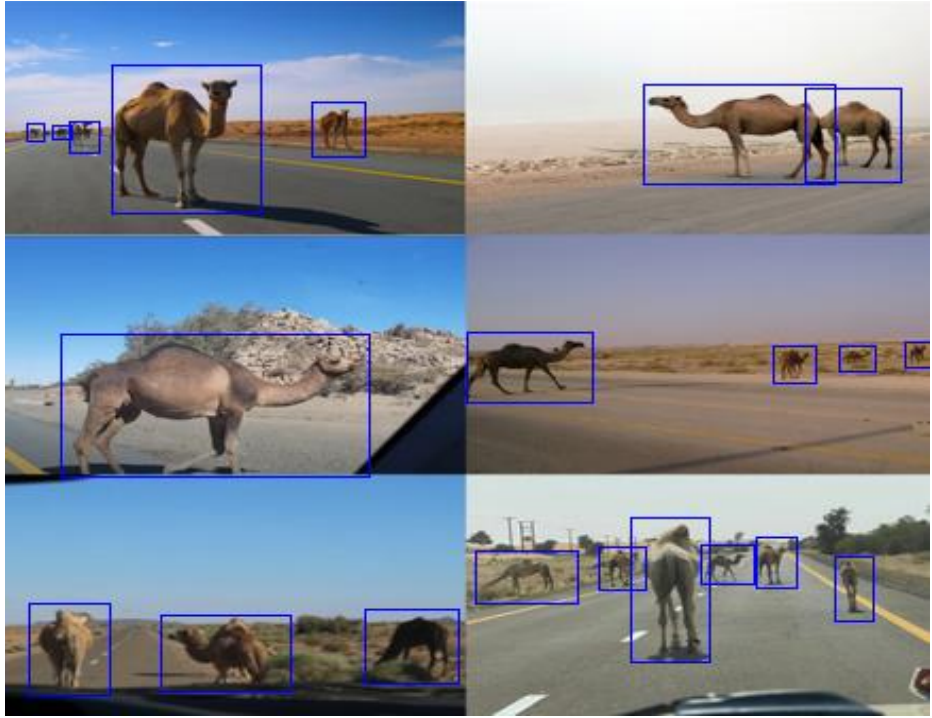

Figure 4. Example images and annotations from the dataset

It is important that each image is accompanied by an annotation file containing the coordinates of the object of interest. This helps to both speed up and optimize the training process. By having these annotations, the model is able to focus only on the relevant features within the given coordinates, disregarding the majority of the image.

## 4. RESULTS AND DISCUSSION

In the experiments, five pre-trained models were used: CenterNet, EfficientDet, Faster R-CNN, SSD, and YOLOv8. This section will first describe how the performance of machine learning models is generally evaluated, followed by a discussion of how object detection models are evaluated specifically. Finally, the results of the experiments will be presented.

### 4.1. Evaluation Metrices

Object detection models are uniformly evaluated using accuracy of the detection boxes through mean Average Precision (mAP) and mean Average Recall (AR). Intersection over Union (IoU) (Figure 5), Recall, and Precision, are helper metrics used to obtain the desired mAP measurement.

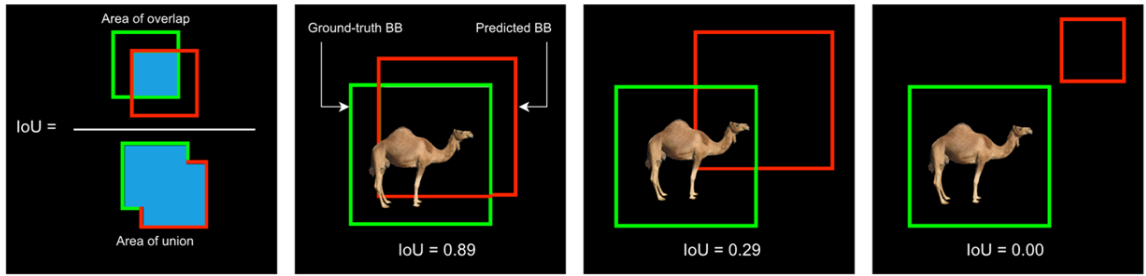

Figure 5. IoU calculation formula. IoU of 0.50:0.95 is considered TP detection.

First, some common evaluation terms and abbreviation widely used in ML model evaluation are the following: True positive (TP), Correct detection made by the model. True Negative (TN), No detection where/when none needed by model. False Positive (FP), incorrect detection made by the model. False Negative (FN), missed detection by model.

To reach the desired performance metrics, the procedure required of following the four equations 1-4 are used to find the mAP:

1) $Precision = \frac{TP}{TP+FP}$
2) $Recall = \frac{TP}{TP+FN}$
3) $Average\ Precision = \int_0^1 p_{(r)}\,dr$
4) $mAP = \frac{1}{N}\sum_{i=1}^{N} AP$

Note, *P(r)* being the curve of plotting Precision-Recall, and *N* as number of classes.

## 4.2. Results

Most of the object detection models used in the experiments were obtained from the TensorFlow 2 Detection Model Zoo repository [25]. These models were pre-trained on the well-known COCO 2017 dataset. The YOLOv8 model was obtained from the ultralytics repository [26].

The models can be adapted to custom datasets through the process of few-shot training. Few-shot learning is a type of machine learning where a model is trained on a limited number of examples and can then generalize to new examples. This is useful in cases where it is challenging or expensive to acquire a large amount of labelled training data, as the model can learn to classify new examples using only a few examples as support. Through transfer learning, we can fine-tune the models to detect camels or any other desired object.

The models utilized for the experiments were: CenterNet, EfficientDet, Faster R-CNN, SSD, and YOLOv8. All models were trained on the same computer using a NVIDIA GeForce GTX 1080 GPU. The accuracy of the models can be viewed in the table below (Table 1) and also visualized in the figure (Figure 2).

Table 1. Model performance comparison

| Model | mAP | | | AR | TrainingTime (m) |
|---|---|---|---|---|---|
| | *IoU=0.50* | *IoU=0.75* | *IoU=0.50:0.95* | *IoU=0.50:0.95* | |
| *CenterNet* | *83.4* | *62.7* | *58* | *33.1* | *35.5* |
| *EfficientDet* | *81.7* | *55.4* | *52.6* | *31.8* | *47.4* |
| *Faster R-CNN* | *80.4* | *62.2* | *52.5* | *31.1* | *85.8* |
| *SSD* | *74.8* | *60.2* | *47.9* | *29* | *52.2* |
| *YOLOv8* | *88.1* | *76.1* | *63.2* | *35.3* | *31.6* |

*All models were trained for 10,000 steps (100 epochs).

YOLOv8 showed superiority in terms of accuracy and efficiency compared to the other models. This result is expected as YOLOv8 is considered the current state-of-the-art object detection algorithm, delivering the highest accuracy while also having a reasonable training time. CenterNet performed well, coming in second place with good general detection rates and training time, although it was less precise compared to YOLOv8. On the other hand, Faster R-CNN, as a two-stage algorithm, had a slower training process and was unable to outperform its one-stage algorithm competitors. It is therefore clear that YOLOv8 is the preferred algorithm for future development of this work.

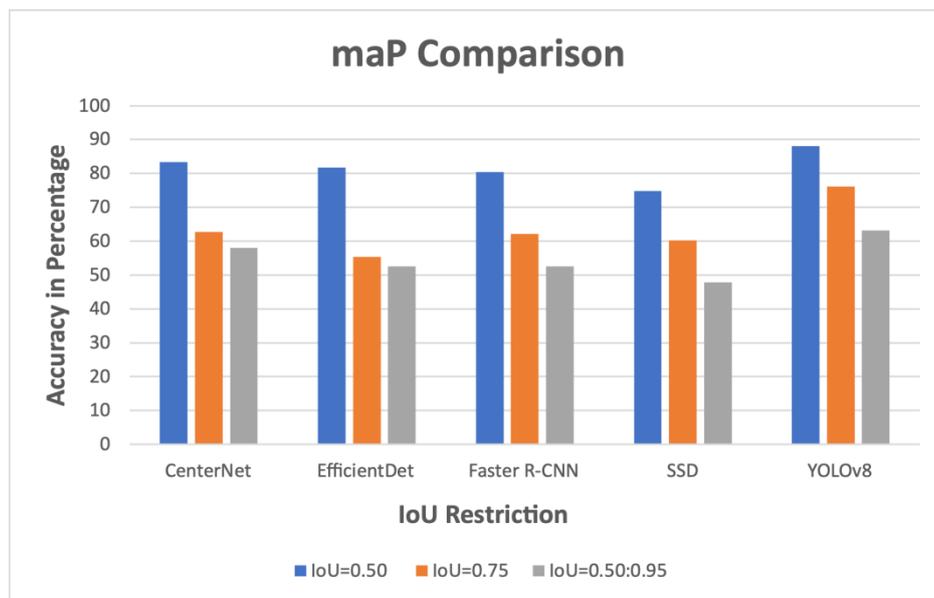

Figure 5. Clustered column graph of models' performances

## 5. CONCLUSION

In conclusion, WVC are a global issue that have a significant socio-economic impact, resulting in billions of dollars in property damage and fatalities. In Saudi Arabia, CVC pose a particularly deadly threat due to the large size of camels, leading to a higher fatality rate compared to other animals. This problem is expected to escalate with the growth of the human population and increasing urbanization. Despite efforts to address CVC, current solutions have not been effective and are often costly. With advancements in technology, the implementation of AI and computer vision in autonomous systems, such as warning systems, has the potential to significantly increase road safety. These systems have already demonstrated success in areas such as enforcing speed and texting and driving laws through the use of cameras.

Based on the findings, significance of the problem, collection of the dataset, and promising results obtained, there is a vision to continue this work by developing a fully autonomous system to make rural areas safer. The system will trigger and disarm alarms based solely on the presence of camels (or other wildlife) near roads. By deploying the developed algorithms on edge devices, the system will be distributed as standalone devices along rural roads, offering a cost-effective solution to traditional fencing, requiring minimal maintenance and providing more effective caution signals than reflective signs.

## 6. THE DATASET

Part of the contribution of this research is to provide a novel type of data that does not exist. A dataset of clean format images and annotated in two styles: Pascal, and YOLO format.

https://www.kaggle.com/datasets/khalidalnujaidi/images-of-camels-annotated-for-object-detection

**AUTHORS**

**Khalid AlNujaidi**. An undergraduate student studying computer science at Prince Mohamed bin Fahad University. Completed a research assistant internship, then continued out of curiosity to learn the field of machine learning and artificial intelligence.

**GhadahAlHabib.** A senior software engineering student at Prince Mohammed bin Fahd University. After completing her bachelor's degree, she will be pursuing a master's degree in artificial intelligence at King Fahd University for Petroleum and Minerals. She has conducted research on Seismic Structures classification using novel features from seismic images. She is currently developing an AI-based innovative education system for the blind. The system is composed of a ring-like device that detects braille and converts it into speech.

**AbdulazizAlOdhieb.** A computer engineering graduate from prince Mohamad bin Fahd University. Abdelaziz completed his internship last summer at Schlumberger as a data scientist, and has been invited back to continue his work with the corporation.